\documentclass{article}

\usepackage[preprint]{neurips_2026}

\usepackage[utf8]{inputenc}
\usepackage[T1]{fontenc}
\usepackage{algorithm}
\usepackage{algpseudocode}
\usepackage{amsmath}
\usepackage{amsfonts}
\usepackage{amssymb}
\usepackage{booktabs}
\usepackage{courier}
\usepackage{enumitem}
\usepackage{float}
\usepackage{graphicx}
\usepackage{microtype}
\usepackage{tikz}
\usepackage{xcolor}
\usepackage{hyperref}
\usepackage{url}

\title{Training Communication-Efficient Mixture-of-Experts \\
Language Models with Layer Re-Configuration}

\author{%
  Simeng Sun, Roger Waleffe\\
  \texttt{\{simengs,rwaleffe\}@nvidia.com}
}

\newcommand{\GBS}{\mathrm{GBS}}

\begin{document}

\maketitle

\begin{abstract}

When training Mixture-of-Experts (MoE) language models with expert parallelism, all-to-all token dispatch and combine collectives can consume a substantial fraction of end-to-end training time. In this work, we study communication-efficient MoE models (CE-MoE), in which we adopt a heterogeneous layer pattern that decouples token-mixing and channel-mixing depth. Compared to conventional models which interleave MoE layers after each token-mixing layer (e.g., attention, Mamba-2), CE-MoE models concentrate expert capacity in a select few routed MoE layers, while maintaining depth by adding additional token-mixing and dense-FFN layers. Across a scaling ladder from 2B to 31.5B total parameters, under matched total and activated parameters, CE-MoE models consistently reduce training cost while matching validation loss and downstream benchmarks with full-MoE baselines. At the 31.5B scale, CE-MoE uses 33.3\% fewer GPU-hours while improving average downstream score and inference throughput.

\end{abstract}

\section{Introduction}
\label{sec:intro}

When training large Mixture-of-Experts (MoE) models with expert parallelism (EP), routing tokens to experts requires cross-rank communication rather than only local computation. In the forward pass, each MoE layer uses all-to-all collectives to dispatch token activations to the ranks hosting their selected experts and combine the resulting expert outputs back on the original token ranks. The backward pass uses all-to-all collectives for gradients along the same routing pattern. Together, these EP collectives can involve large communication volume and, in practice, consume a substantial fraction of end-to-end training time~\citep{yan2026megatroncore}, even with moderate routed top-$K$. Specifically, with dropless routing and no capacity padding, the expert-parallel communication volume over one training step (forward+backward) $V_{\mathrm{step}}$ scales as
\begin{equation}
  V_{\mathrm{step}}
  \propto
  L_E \cdot S \cdot \GBS \cdot K
  \left(4 D b_x + 2 b_p\right),
  \label{eq:intro-v-step}
\end{equation}
where 
$L_E$ is the number of MoE layers, 
$S$ is the sequence length, 
$\GBS$ is the global batch size, 
$K$ is the routed top-$K$, 
$D$ is the dispatch hidden dimension, 
$b_x$, $b_p$ are the number of bytes per hidden element (e.g. BF16), and per routed probability (FP32) respectively. 

System-level optimizations such as collective algorithms, computation-communication overlap, and kernel efficiency  can reduce exposed EP communication cost. Our focus is orthogonal and complementary: reducing the total volume of moved data in the first place, so that system-level optimizations can compound the gains. The communication volume as modeled in Eq.~\ref{eq:intro-v-step} exposes multiple levers for reducing total communication payload:
\begin{itemize}[leftmargin=*]
\small
  \item \textbf{Precision} ($b_x$, $b_p$): use lower-precision for communication~\citep{nvidia2025nvfp4}.
  \item \textbf{Dispatch dimension} ($D$): adopt LatentMoE-style~\citep{elango2026latentmoe} projections that dispatch routed activations through a lower-dimensional latent space.
  \item \textbf{Routing width} ($K$): reduce routed top-$K$ so each token is dispatched to fewer experts.
  \item \textbf{MoE depth} ($L_E$): reduce the number of MoE layers.
\end{itemize}

We study MoE models with hybrid token-mixer blocks~\citep{lieber2024jamba,nvidia2025nemotron3,nvidia2026nemotron3super}, where token-mixers are either full softmax attention
or linear recurrent blocks.
Rather than regularly pairing token-mixing blocks with routed MoE layers, we substantially reduce $L_E$, while keeping $K$ and $D$ in a controlled range so that total EP all-to-all volume is reduced.
We refer to such architectures as communication-efficient MoE (CE-MoE). CE-MoE replaces the removed MoE layers with additional token-mixing and dense-FFN blocks, and compensates for fewer MoE layers with more and wider experts per MoE layer.
The resulting token-mixing-heavy layout is roughly parameter matched but more heterogeneous than the regular baseline layer layout. The consecutive token-mixing layers as present in the CE-MoE layout can trigger rank collapse: token representations concentrate into a few dominant directions and develop large norms that destabilize training. We empirically show that linear recurrent blocks are less prone to this instability than softmax-attention token mixers.
Finally, we provide a recipe for constructing CE-MoE models that match the downstream performance of parameter-matched baseline hybrid-MoE models while reducing end-to-end training cost.

The rest of the report is organized as follows. 
\begin{itemize}[leftmargin=*]

\item \textbf{Section~\ref{sec:overview}} situates CE-MoE among approaches for reducing EP cost and motivates the heterogeneous layer layout.
\item \textbf{Section~\ref{sec:rank-collapse-analysis}} examines training instability in consecutive token mixers and shows Mamba-2 is more stable than softmax attention token-mixer.
\item \textbf{Section~\ref{sec:reduced-moe-approach}} provides a recipe for constructing a CE-MoE model using a running example.
\item \textbf{Section~\ref{sec:empirical-results}} evaluates CE-MoE against parameter-matched hybrid-MoE baselines on training cost, downstream quality, inference throughput, and residual-stream behavior.
\item \textbf{Section~\ref{sec:limitations-summary}} summarizes the main findings and discusses limitations.
\end{itemize}

\section{Overview}
\label{sec:overview}

\paragraph{Related work.}
Related work on reducing the cost of large-scale MoE training spans several levels. At the kernel level, SonicMoE~\citep{guo2025sonicmoe} improves MoE efficiency through tile-aware optimizations.
At the system and framework level, DeepEP~\citep{deepep} provides high-throughput and low-latency all-to-all kernels, while Megatron-Core~\citep{yan2026megatroncore} and MegaScale-MoE~\citep{jin2026megascale} provide scalable systems stacks for large-scale MoE training. 
At the architecture-system boundary, shortcut-connected MoE~\citep{cai2024shortcut} changes the dependency structure so that EP communication can overlap with computation in the paired attention stream. Equation~\ref{eq:intro-v-step} shows that, the logical communication volume scales with dispatch hidden dimension $D$. LatentMoE~\citep{elango2026latentmoe} reduces the effective dispatch dimension $D$, thereby reducing all-to-all traffic. Multi-Head LatentMoE with head parallel~\citep{cui2026multiheadlatentmoe} changes the architecture and parallelism so communication no longer scales linearly with the number of activated experts.

\begin{figure}[H]
  \centering
  \resizebox{\linewidth}{!}{%
  \begin{tikzpicture}[
    >=stealth,
    font=\sffamily\scriptsize,
    block/.style={draw, rounded corners=2pt, minimum height=0.78cm,
      minimum width=1.25cm, align=center, inner sep=2pt},
    attention/.style={fill=blue!12, draw=blue!60!black},
    mamba/.style={fill=violet!13, draw=violet!70!black},
    dense/.style={fill=green!14, draw=green!50!black},
    moe/.style={fill=orange!18, draw=orange!70!black},
    edge/.style={->, draw=black!55, line width=0.35pt},
    legend/.style={draw, rounded corners=1.5pt, minimum height=0.26cm,
      minimum width=0.42cm, inner sep=0pt}
  ]
    \node[font=\sffamily\bfseries, anchor=east] at (0.0,1.85)
      {MoE baseline};
    \node[block, attention] (d1) at (1.05,1.85) {Attention};
    \node[block, moe, minimum width=1.48cm, minimum height=0.88cm]
      (d2) at (2.68,1.85) {MoE};
    \node[block, attention] (d3) at (4.32,1.85) {Attention};
    \node[block, moe, minimum width=1.48cm, minimum height=0.88cm]
      (d4) at (5.87,1.85) {MoE};
    \node[block, attention] (d5) at (7.42,1.85) {Attention};
    \node[block, moe, minimum width=1.48cm, minimum height=0.88cm]
      (d6) at (9.05,1.85) {MoE};
    \node[block, attention] (d7) at (10.68,1.85) {Attention};
    \node[block, moe, minimum width=1.48cm, minimum height=0.88cm]
      (d8) at (12.23,1.85) {MoE};
    \coordinate (dnext) at (13.18,1.85);
    \node[anchor=west, font=\sffamily\large] at (13.34,1.85) {$\cdots$};
    \draw[edge] (d1) -- (d2);
    \draw[edge] (d2) -- (d3);
    \draw[edge] (d3) -- (d4);
    \draw[edge] (d4) -- (d5);
    \draw[edge] (d5) -- (d6);
    \draw[edge] (d6) -- (d7);
    \draw[edge] (d7) -- (d8);
    \draw[edge] (d8) -- (dnext);

    \node[font=\sffamily\bfseries, anchor=east] at (0.0,0.55)
      {Switch/GShard-style};
    \node[block, attention] (s1) at (1.05,0.55) {Attention};
    \node[block, dense] (s2) at (2.68,0.55) {Dense\\FFN};
    \node[block, attention] (s3) at (4.32,0.55) {Attention};
    \node[block, moe, minimum width=1.48cm, minimum height=0.88cm]
      (s4) at (5.87,0.55) {MoE};
    \node[block, attention] (s5) at (7.42,0.55) {Attention};
    \node[block, dense] (s6) at (9.05,0.55) {Dense\\FFN};
    \node[block, attention] (s7) at (10.68,0.55) {Attention};
    \node[block, moe, minimum width=1.48cm, minimum height=0.88cm]
      (s8) at (12.23,0.55) {MoE};
    \coordinate (snext) at (13.18,0.55);
    \node[anchor=west, font=\sffamily\large] at (13.34,0.55) {$\cdots$};
    \draw[edge] (s1) -- (s2);
    \draw[edge] (s2) -- (s3);
    \draw[edge] (s3) -- (s4);
    \draw[edge] (s4) -- (s5);
    \draw[edge] (s5) -- (s6);
    \draw[edge] (s6) -- (s7);
    \draw[edge] (s7) -- (s8);
    \draw[edge] (s8) -- (snext);

    \node[font=\sffamily\bfseries, anchor=east] at (0.0,-0.75)
      {CE-MoE (Ours)};
    \node[block, mamba, minimum width=1.05cm] (r1) at (1.05,-0.75) {Mamba2};
    \node[block, mamba, minimum width=1.05cm] (r2) at (2.35,-0.75) {Mamba2};
    \node[block, moe, minimum width=1.95cm, minimum height=0.96cm]
      (r3) at (4.05,-0.75) {MoE};
    \node[block, mamba, minimum width=1.05cm] (r4) at (5.75,-0.75) {Mamba2};
    \node[block, mamba, minimum width=1.05cm] (r5) at (7.05,-0.75) {Mamba2};
    \node[block, dense, minimum width=1.12cm] (r6) at (8.35,-0.75) {Dense\\FFN};
    \node[block, attention, minimum width=1.10cm] (r7) at (9.65,-0.75) {Attention};
    \node[block, mamba, minimum width=1.05cm] (r8) at (10.95,-0.75) {Mamba2};
    \node[block, mamba, minimum width=1.05cm] (r9) at (12.25,-0.75) {Mamba2};
    \coordinate (rnext) at (13.18,-0.75);
    \node[anchor=west, font=\sffamily\large] at (13.34,-0.75) {$\cdots$};
    \draw[edge] (r1) -- (r2);
    \draw[edge] (r2) -- (r3);
    \draw[edge] (r3) -- (r4);
    \draw[edge] (r4) -- (r5);
    \draw[edge] (r5) -- (r6);
    \draw[edge] (r6) -- (r7);
    \draw[edge] (r7) -- (r8);
    \draw[edge] (r8) -- (r9);
    \draw[edge] (r9) -- (rnext);

    \node[legend, attention] at (0.85,-1.85) {};
    \node[anchor=west] at (1.12,-1.85) {token-mixing (attention)};
    \node[legend, mamba] at (4.42,-1.85) {};
    \node[anchor=west] at (4.69,-1.85) {token-mixing (Mamba2)};
    \node[legend, dense] at (7.72,-1.85) {};
    \node[anchor=west] at (7.99,-1.85) {channel-mixing (dense)};
    \node[legend, moe] at (11.55,-1.85) {};
    \node[anchor=west] at (11.82,-1.85) {channel-mixing (MoE)};
  \end{tikzpicture}%
  }
  \caption{MoE baseline and Switch/GShard-style sparse model preserve a regular token-mixing/channel-mixing alternation.
  CE-MoE uses a token-mixing-heavy heterogeneous layout.}
  \label{fig:layer-layout-overview}
\end{figure}
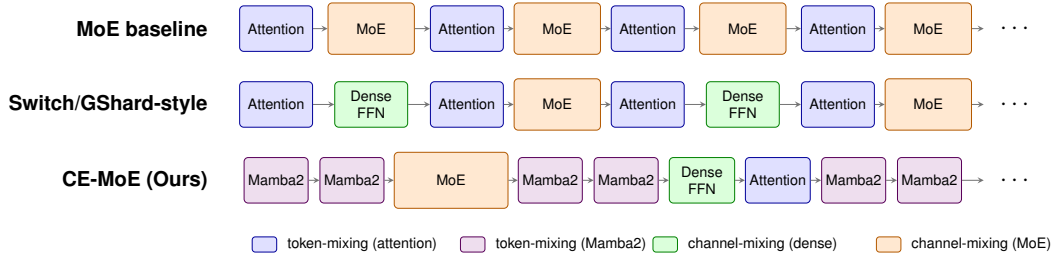

\paragraph{Reducing MoE depth $L_E$.}
Existing large-scale MoE architectures reduce routed MoE depth through interleaving MoE and dense FFN blocks~\citep{lepikhin2020gshard,fedus2022switch,zoph2022stmoe,mai2026thinking1}, e.g., Switch Transformer places MoE at every other FFN layers. These designs nevertheless preserve the regular alternation between token-mixing and channel-mixing blocks in standard Transformers (e.g., \texttt{attn} $\rightarrow$ \texttt{dense mlp} $\rightarrow$ \texttt{attn} $\rightarrow$ \texttt{moe}). In contrast, we reduce MoE depth more aggressively and decouple token-mixing from channel-mixing counts, producing more heterogeneous layer layouts that depart from the standard Transformer block design, as shown in Figure~\ref{fig:layer-layout-overview}.

The heterogeneity in our communication-efficient hybrid-MoE (CE-MoE) is a direct consequence of adopting a skewed
distribution between token-mixing and channel-mixing layers. Compared to standard Transformer with 1:1 token-mixing to channel-mixing ratio, CE-MoE allocates more layers for exchanging information along sequence dimension, assigning four token-mixing layers for every MoE layer or more than two token-mixing layers for every channel-mixing layer. 

With the token-mixing-heavy layer composition, we use a simple \emph{phased-greedy procedure} to construct the CE-MoE layer pattern, spacing attention and channel-mixing blocks along a spine with linear recurrent blocks while permitting consecutive token-mixing blocks. This design raises the main concern addressed in the next section with a toy setting: whether consecutive
token-mixing layers remain expressive and stable to train. In later sections, we empirically verify at larger scale that the resulting pattern recovers the downstream quality of parameter-matched hybrid-MoE baselines while reducing end-to-end training cost.

\section{Rank-collapse risk}
\label{sec:rank-collapse-analysis}

\begin{figure}[!t]
  \centering
  \includegraphics[width=0.5\linewidth]{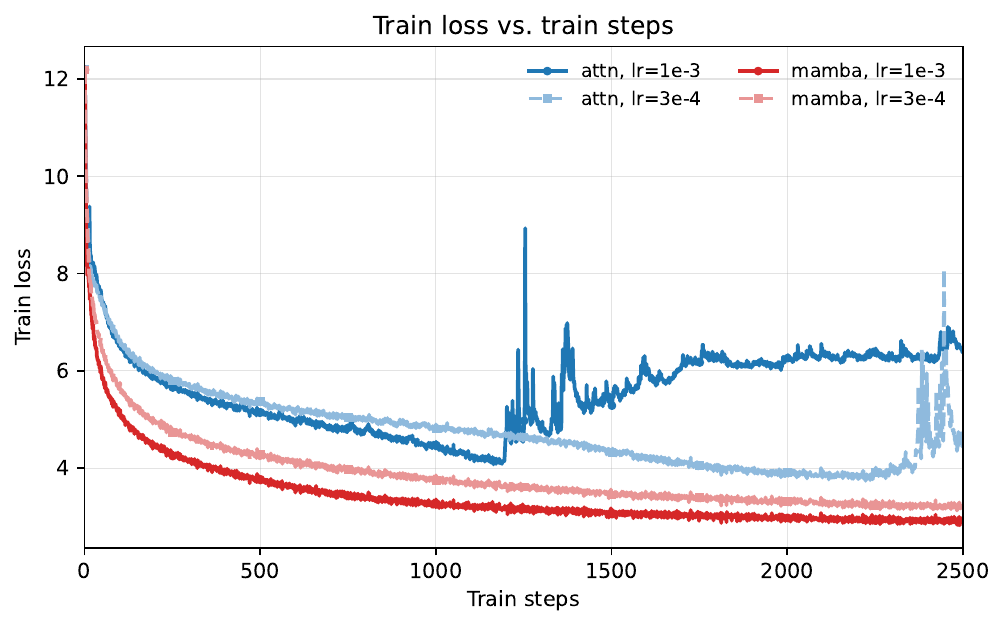}
  \caption{Training loss for tiny models of $\sim$370M parameters with seven consecutive token-mixing
  layers. The attention variants are less stable, while the Mamba-2 variants train
  smoothly at both learning rates.}
  \label{fig:rank-collapse-train-loss}
\end{figure}

The token-mixing-heavy layer composition in CE-MoE introduces adjacent token-mixing blocks. This raises concerns about training stability, as existing work shows that pure self-attention stacks without MLPs or skip connections can converge doubly exponentially to rank~1 \citep{dong2021attention} thus losing expressivity. We empirically show that self-attention stacks are unstable to train even with skip connections. Figure~\ref{fig:rank-collapse-train-loss} compares $\sim$370M-parameter models with seven consecutive token-mixing layers (self-attention or Mamba2~\citep{dao2024transformers}) at two learning rates. The attention variants show loss spikes and eventually diverge, whereas the Mamba2 variants converge smoothly; the two architectures are matched in per-layer parameter count.

To quantify token diversity and rank changes, we measure layerwise statistics on post-residual representations $X \in \mathbb{R}^{N_{\mathrm{tok}} \times D}$, where $N_{\mathrm{tok}}$ is the number of sampled evaluation tokens and $D{=}1024$ is the hidden dimension. Let $\sigma_i$ denote the singular values of $X$, $\lVert \cdot \rVert_F$ the Frobenius norm, $p_i = \sigma_i^2 / \sum_j \sigma_j^2$ the normalized spectral distribution, and $H(p)$ its Shannon entropy. We report stable rank
$
  \mbox{\footnotesize $r_{\mathrm{stable}}(X) = \lVert X \rVert_F^2 / \sigma_1^2$},
$
and effective rank
$
  \mbox{\footnotesize $r_{\mathrm{effective}}(X) = \exp(H(p))$.}
$
When either metric approaches~1, most residual-stream energy lies in a single singular direction. Token representations become nearly collinear, leaving token-specific information to be carried mainly by scalar magnitudes or small residual components.

\begin{figure}[!t]
  \centering
  \begin{minipage}{0.49\linewidth}
    \centering
    \includegraphics[width=\linewidth]{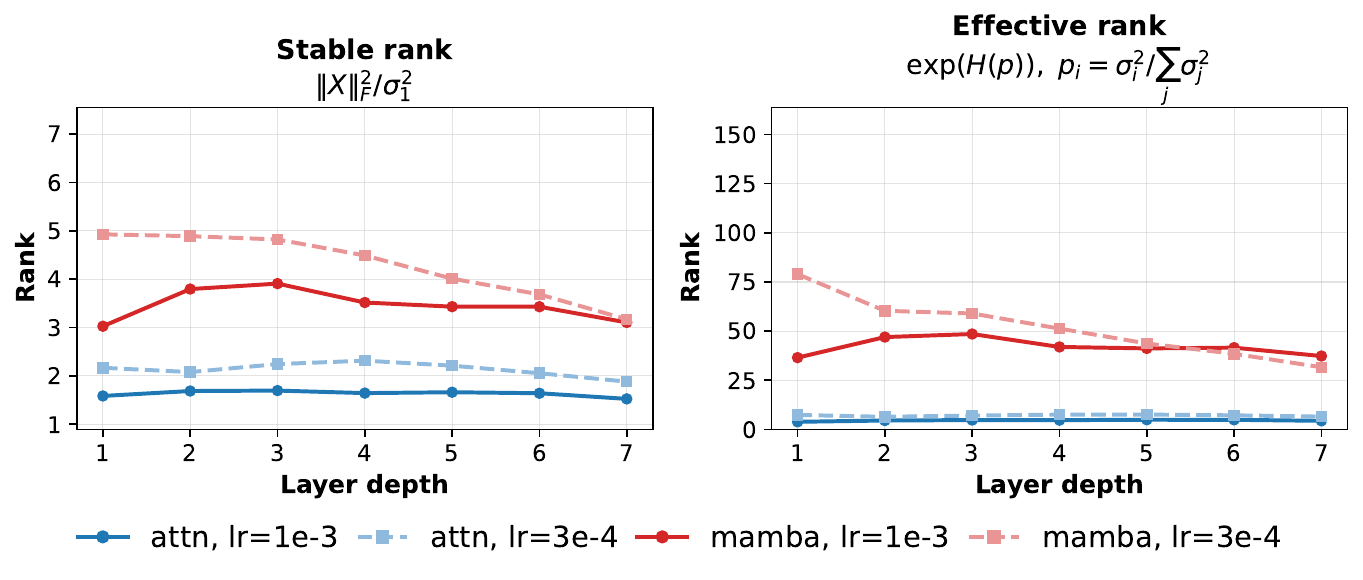}
    \vspace{-0.5em}
    {\small (a) Step 125}
  \end{minipage}
  \hfill
  \begin{minipage}{0.49\linewidth}
    \centering
    \includegraphics[width=\linewidth]{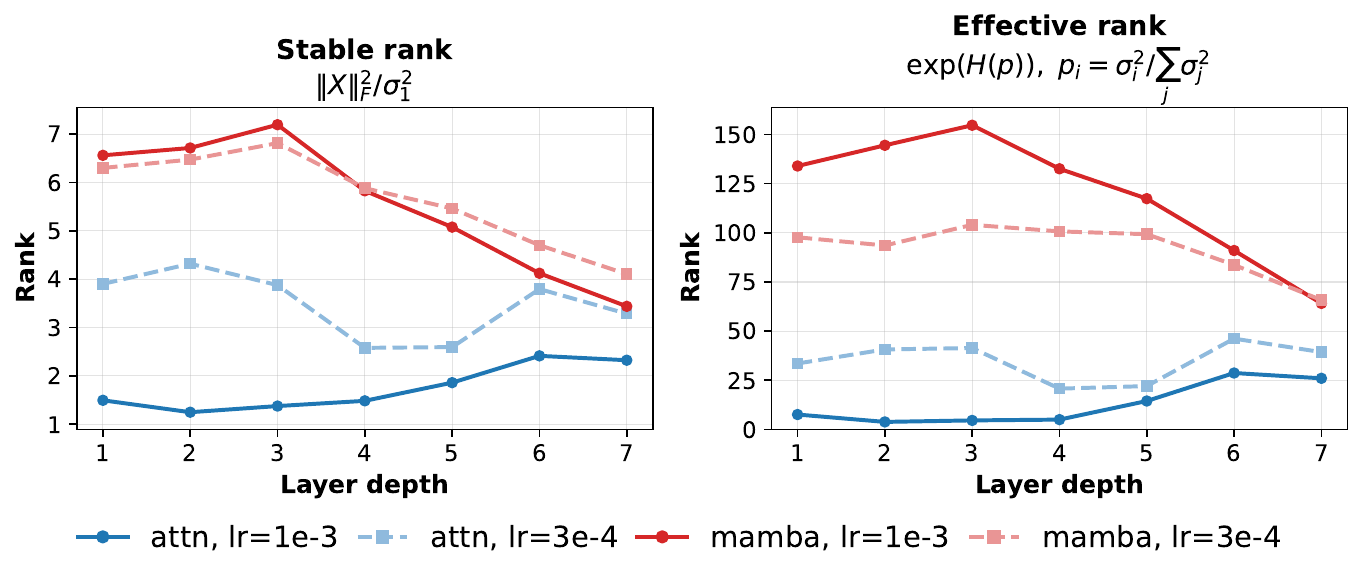}
    \vspace{-0.5em}
    {\small (b) Step 1000}
  \end{minipage}
  \caption{Layerwise stable rank and effective rank near the beginning of training and at step 1000, before divergence of the self-attention runs. Consecutive self-attention exhibits severe low-rank bottleneck, while Mamba2 maintains higher ranks despite decreasing with depth.}
  \label{fig:rank-collapse-rank-stats}
\end{figure}

Figure~\ref{fig:rank-collapse-rank-stats} shows that the self-attention networks have much lower stable and effective rank than Mamba-2, with the strongest collapse in early layers. At the higher learning rate, the early self-attention layers reach stable rank close to~1 and single-digit effective rank, indicating that most residual-stream energy lies in only a few directions despite $D{=}1024$. The high-learning-rate self-attention network also develops per-token $\ell_2$ norms that are orders of magnitude larger than the Mamba-2 for all layers, suggesting that, rather than by diverse directions, token differences are increasingly carried by magnitude, which serves as a major source of instability.

These results suggest that consecutive token-mixing blocks are viable when the token mixer maintains healthy rank under repeated stacking. In the toy experiments, the linear recurrent block (Mamba-2) demonstrates better stability than self-attention and is therefore a better candidate for the main token-mixing path in a heterogeneous layout. Other mechanisms, such as gated attention~\citep{qiu2025gatedattention} and normalized GPT~\citep{loshchilov2024ngpt} may help stablize pure self-attention networks by introducing new directions or controlling residual-stream norms, however they are outside the scope of this report. For the remaining sections, we focus on hybrid models (softmax attention and linear attention) where the linear recurrent block provides the architectural flexibility needed to construct token-mixing-heavy CE-MoE layer patterns.

\section{Communication-Efficient Hybrid-MoE}
\label{sec:reduced-moe-approach}

The goal is to design a communication-efficient hybrid-MoE (CE-MoE) architecture that preserves the downstream quality of a baseline hybrid-MoE model while using far fewer routed MoE layers. We build CE-MoE in three steps: (1) determine the layer composition, (2) generate the layer pattern with a phased-greedy procedure, and (3) adjust model shape so the transformed model remains comparable to the baseline under the target parameter budget.

\subsection{Determine the layer composition}
Standard Transformer architecture treats the combination of attention and dense FFN (or MoE) as a single layer and therefore implicitly enforces a $1{:}1$ token-mixing-to-channel-mixing ratio. CE-MoE relaxes the 1:1 ratio by reducing MoE layers and reallocating depth to additional token-mixing layers together with a moderate number of dense FFN blocks.

Let the baseline composition be $(a^\star,b^\star,c^\star)$ for Mamba, MoE, and attention blocks, with $L=a^\star+b^\star+c^\star$. Let CE-MoE use $(a,b,c,d)$ for Mamba, MoE, attention, and dense FFN blocks while preserving the same total $L=a+b+c+d$. 
CE-MoE sets block counts by targeting an approximate token-mixing-to-MoE ratio $\alpha$\footnote{Preliminary experiments indicate that values in $[4,5]$ preserve downstream quality before performance degrades noticeably.} and keeping $c=c^\star$ to control long-context attention cost. With $d=b$,\footnote{Preliminary ablations showed no clear trend between model quality and the dense-to-MoE ratio; we therefore take $d=b$ for simplicity.} this gives $b=\left\lfloor \frac{L}{\alpha+2} \right\rfloor$, with the remaining blocks assigned to Mamba.

As a concrete example, let $q$ denote the number of expert-parallel all-to-all invocations per rank in an EP group for one forward-plus-backward microbatch through one routed MoE layer.\footnote{Under Megatron-LM's \texttt{alltoall} dispatcher, $q=6$: hidden-state dispatch, routed-probability dispatch, and hidden-state combine in the forward pass, plus the inverse all-to-alls in the backward pass.} For a Nemotron-3 Nano baseline composition $(a^\star,b^\star,c^\star)=(23,23,6)$ (total $L=52$)~\citep{nvidia2025nemotron3}, the layer composition for a corresponding CE-MoE model, when we target an approximate token-mixing-to-MoE ratio $\alpha=4$, is thus $b=d=\left\lfloor 52/(4+2)\right\rfloor=8$, and $a=30$. 
The baseline alternating layout, the every-other-layer (Switch-style) variant, and CE-MoE therefore incur $23q$, $12q$, and $8q$ MoE all-to-all calls per rank and microbatch within an EP group, respectively. 
Relative to the baseline, every-other-layer (Switch-style) cuts communication by $47.8\%$ and CE-MoE by $65.2\%$. The layer-count difference also governs MoE all-to-all volume up to routed top-$K$ and dispatch dimension (Equation~\ref{eq:intro-v-step}); as an illustrative case, if both $L_E$-reduced variants use $1.5K_{\mathrm{base}}$ and identical dispatch dimension, Switch-style lowers volume by $\simeq 21.7\%$ and CE-MoE by $\simeq 47.8\%$.

\subsection{Generate the layer pattern}

The composition fixes the multiset of blocks but not their order. CE-MoE assigns layer order with a phased-greedy procedure in Algorithm~\ref{alg:ce-moe-layer-pattern}, where \texttt{M}, \texttt{E}, \texttt{*}, and \texttt{-} denote Mamba, MoE, attention, and dense FFN blocks, respectively.

The algorithm spreads self-attention and channel-mixing blocks across a Mamba-2 spine while avoiding poor local orderings. It runs in four sequential phases. Phases 1--3 populate three types of virtual blocks: $A$ (attention-bearing blocks), $N$ (regular non-attention blocks), and $R$ (residue blocks). 
We use the tuple $(m,\mathbf{1}_{e},\mathbf{1}_{a},\mathbf{1}_{d})$ to denote a virtual block, where $m$ is the number of Mamba-2 blocks stored in a virtual block and $\mathbf{1}_{e},\mathbf{1}_{a},\mathbf{1}_{d}$ indicate whether MoE, attention, and dense FFN blocks are present. The algorithm first computes a nominal Mamba budget $\ell=\lceil n_M/n_{\mathrm{vb}}\rceil$ for regular virtual blocks. Phases 1--2 greedily create attention-bearing and non-attention blocks with $m=\ell$; Phase 3 stores any leftover layers in residue blocks $R$, whose Mamba count may be smaller. In Phase 4, $\mathrm{Distribute}(A,N)$ treats each virtual block atomically and places $N$ blocks so they evenly partition the sequence of $A$ blocks, breaking ties from left to right. Blocks in $R$ are appended at the end. 

Within each virtual block, the same spacing principle is applied: channel-mixing blocks are distributed along the Mamba spine, attention (if present) sits at the token-mixing tail, and the block ends with channel mixing so the next virtual block can begin with token mixing.

\begin{algorithm}[t]
\caption{Generate CE-MoE layer pattern}
\label{alg:ce-moe-layer-pattern}
\small
\begin{algorithmic}[1]
\Require Composition $(a,b,c,d)$ over (Mamba, MoE, attention, dense FFN), $b \geq c$
\Ensure Layer pattern over \texttt{M}, \texttt{E}, \texttt{*}, and \texttt{-}
\State $(n_M,n_E,n_A,n_D) \gets (a,b,c,d)$; \quad $n_{\mathrm{vb}} \gets \max(n_E,n_A,n_D,1)$; \quad $\ell \gets \lceil n_M/n_{\mathrm{vb}}\rceil$
\State \Comment{\textcolor{lightgray}{A virtual-block tuple $(m,\mathbf{1}_{e},\mathbf{1}_{a},\mathbf{1}_{d})$ uses the indicator variables defined above; \(\mathbf{1}[\cdot]\) is the indicator function; \(\|\) denotes string or list concatenation.}}
\State $A,N,R \gets [\,],[\,],[\,]$
\While{$n_A > 0$ \textbf{and} $n_M \ge \ell$} \Comment{\textcolor{lightgray}{Phase 1: emit attention-bearing virtual blocks}}
    \State $(\mathbf{1}_{e},\mathbf{1}_{d}) \gets (\mathbf{1}[n_E>0],\mathbf{1}[n_D>0])$
    \State $A.\mathrm{append}((\ell,\mathbf{1}_{e},1,\mathbf{1}_{d}))$
    \State $(n_M,n_E,n_A,n_D) \gets (n_M-\ell,n_E-\mathbf{1}_{e},n_A-1,n_D-\mathbf{1}_{d})$
\EndWhile
\While{$n_M \ge \ell$ \textbf{and} $n_E > 0$ \textbf{and} $n_D > 0$} \Comment{\textcolor{lightgray}{Phase 2: emit regular non-attention virtual blocks}}
    \State $N.\mathrm{append}((\ell,1,0,1))$
    \State $(n_M,n_E,n_D) \gets (n_M-\ell,n_E-1,n_D-1)$
\EndWhile
\If{$n_M+n_E+n_A+n_D > 0$} \Comment{\textcolor{lightgray}{Phase 3: collect residual counts into shortened tail block(s)}}
    \State append residue virtual blocks with $m>0$ covering the remaining counts to $R$
\EndIf
\State $\mathit{blocks} \gets \mathrm{Distribute}(A,N) \,\|\, R$ \Comment{\textcolor{lightgray}{Phase 4: place $N$ to evenly partition $A$, breaking ties left-to-right}}
\State $P \gets$ empty string
\ForAll{virtual blocks $(m,\mathbf{1}_{e},\mathbf{1}_{a},\mathbf{1}_{d}) \in \mathit{blocks}$}
    \State $s \gets \mathrm{Insert}(\mathtt{M}^{m}, \mathtt{E}, \mathbf{1}_{e}) \,\|\, \mathtt{*}^{\mathbf{1}_{a}} \,\|\, \mathtt{-}^{\mathbf{1}_{d}}$ \Comment{\textcolor{lightgray}{\(\mathrm{Insert}\) returns \(\mathtt{M}^m\) if \(\mathbf{1}_e=0\), else adds \(\mathtt{E}\) after \(\lfloor m/2\rfloor\) Mamba}}
    \State $P \gets P \,\|\, s$
\EndFor
\State \Return $P$
\end{algorithmic}
\end{algorithm}

For the running example of CE-MoE composition $(a,b,c,d)=(30,8,6,8)$, the procedure yields $n_{\mathrm{vb}}=8$ and $\ell=4$. After greedy allocation, $|A|=6$, $|N|=1$, and $|R|=1$; $\mathrm{Distribute}$ then produces the global virtual-block order \texttt{AAANAAAR}:
\begin{center}
    \small
\texttt{[MMEMM*-][MMEMM*-][MMEMM*-](MMEMM-)[MMEMM*-][MMEMM*-][MMEMM*-](MEM-)}.
\end{center}

We run a small layer-order ablation to show the effectiveness of the phased-greedy procedure. At 2B parameters and 140B tokens, we compare the phased-greedy CE-MoE pattern against 21 constrained-random layer patterns that share the same block composition. The constrained-random layer patterns are generated while satisfying three constraints to exclude apparently bad layer patterns: (1) it begins with Mamba and ends with dense FFN, (2) no consecutive channel-mixing blocks (\texttt{EE}, \texttt{E-}, \texttt{-E}, \texttt{---}), and (3) self-attention blocks are placed at least three positions apart. This yields informed random baselines over plausible orderings rather than unconstrained permutations.

\begin{figure}[!t]
  \centering
  \includegraphics[width=0.6\linewidth]{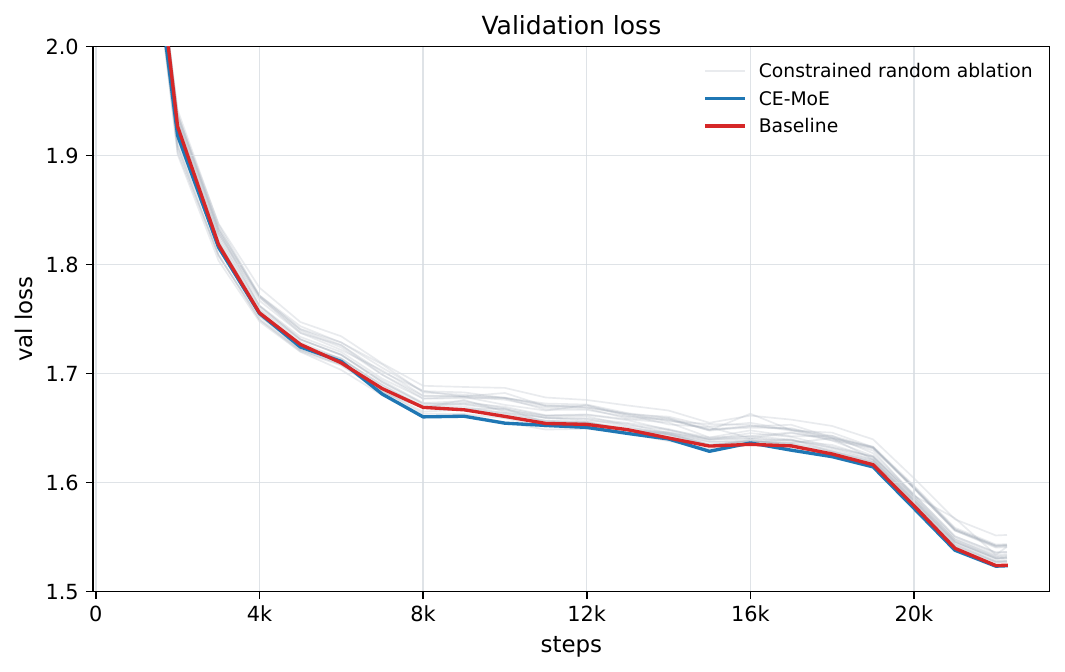}
  \caption{Validation loss for the phased-greedy CE-MoE pattern, the matched baseline, and 21 constrained-random layer patterns at 2B total parameters and 140B tokens.}
  \label{fig:phase1-arch-branch-val-loss}
\end{figure}

Figure~\ref{fig:phase1-arch-branch-val-loss} shows that phased-greedy CE-MoE matches the baseline and outperforms all constrained-random alternatives in terms of validation loss. At the final step, CE-MoE lies roughly $1.8$ standard deviations below the constrained-random mean, suggesting that layer ordering, and not merely layer composition, carries useful inductive bias.

\subsection{Determine the model shape}

Once the layer pattern is fixed, we search for model shape configuration so CE-MoE and the baseline are compared under matched total and activated parameter budgets. Most width and routing ratios are kept near baseline values, while per-expert MoE-FFN width and expert count are allowed to increase to compensate for fewer routed MoE layers. At fixed hidden size, we grid-search expert count, routed top-$K$, per-expert MoE-FFN width, shared-expert width, and dense-FFN width; candidate values are snapped to hardware-friendly multiples, and expert counts must divide evenly across expert-parallel ranks. Among budget-matched candidates, we select the one with the smallest $L_EK$, a communication-cost proxy from Equation~\ref{eq:intro-v-step} when $D$ and communication precision are fixed.

\begin{table}[!h]
  \caption{Width and routing search space for the 31.5B running example. All ranges are relative to the baseline shape.}
  \label{tab:model-shape-search-space}
  \centering
  \small
  \begin{tabular}{ll}
    \toprule
    \textbf{Parameter} & \textbf{Search range} \\
    \midrule
    Hidden size $H$ & $1.0\times$ (fixed) \\
    Dense-FFN width & $3.0$--$5.0\times$ \\
    MoE-FFN width (per expert) & $1.5$--$4.0\times$ \\
    Shared-expert width & $0.95$--$1.15\times$ \\
    Expert count $|\mathcal{E}|$ & $\{1.25,1.5\}\times$; divisible by EP rank\\
    Routing density $K/|\mathcal{E}|$ & $0.5$--$1.5\times$ \\
    Acceptance criterion & total params $\pm 1\%$; activated params $\pm 1.5\%$ \\
    \bottomrule
  \end{tabular}
\end{table}

In the running 31.5B example, the baseline uses $H{=}2688$, per-expert MoE-FFN width $1856$, shared-expert width $3712$, $|\mathcal{E}|{=}128$, $K{=}6$, and $L_E{=}23$, giving $L_EK{=}138$. The search in Table~\ref{tab:model-shape-search-space} spans $1{,}708{,}200$ candidates, of which $5{,}817$ meet the acceptance criterion. The selected CE-MoE shape uses dense-FFN width $8096$, per-expert MoE-FFN width $3520$, shared-expert width $3520$, $|\mathcal{E}|{=}192$, and $K{=}7$. Relative to the baseline, it changes total and activated parameters by $-0.001\%$ and $-0.257\%$, respectively, while reducing $L_EK$ by $59.4\%$.

\section{Empirical results}
\label{sec:empirical-results}

\subsection{Evaluation axes}
We evaluate CE-MoE along three axes:
(1)~\textbf{Training cost.} Architectures that are more efficient FLOPs-wise may still take longer time to train if they introduce system-level bottlenecks (e.g., communication, routing overhead, or inefficient kernels). We therefore use end-to-end training cost in GPU-hours, rather than theoretical FLOPs, as the primary metric for measuring training cost. For each run, we estimate steady-state GPU-hours by discarding warmup iterations, iteratively removing latency outliers above the current mean plus $1.5\sigma$ (checkpoint saves, restarts, and transient network jitter), and multiplying the filtered mean latency by the total iteration count and world size. The resulting estimate reflects optimistic architecture-level training cost and does not equal the billed cluster GPU-hours.
(2)~\textbf{Model quality.} We report validation loss and downstream performance on 11 standardized language-model benchmarks: MMLU and MMLU-Pro~\citep{hendrycks2021mmlu,wang2024mmlupro}; coding benchmarks HumanEval and MBPP~\citep{chen2021evaluating,austin2021program}; math benchmarks GSM8K, MATH-500, and MATH-Hard (Minerva MATH level-5)~\citep{cobbe2021training,hendrycks2021math,lightman2023lets,lewkowycz2022solving}; and commonsense benchmarks RACE, HellaSwag, WinoGrande, and ARC-Challenge~\citep{lai2017race,zellers2019hellaswag,sakaguchi2020winogrande,clark2018think}.
(3)~\textbf{Inference speed.} We run preliminary static inference benchmarks to understand how reducing the MoE layer count affects inference performance. We report tokens per second at input sequence lengths 2048 and 8192 across various output sequence lengths.

\subsection{Experimental details}

All experiments use hybrid Mamba-2 models, with layer compositions reported as \texttt{M+A+D+E} for Mamba-2, self-attention, dense FFN, and MoE layers. Within each model scale, the baseline and CE-MoE variants use the same hidden size and are matched in similar total and activated parameter budgets following the model shape search in Section~\ref{sec:reduced-moe-approach}. Training runs are executed on H100 HBM3 clusters with 4th-gen NVLink.  We train with BF16 precision at sequence length 8192, global batch size 768 unless specified otherwise. Adam optimizer settings \((\beta_1,\beta_2)=(0.9,0.95)\), weight decay 0.1, gradient clipping at 1.0, and WSD learning-rate schedule. MoE layers use \texttt{alltoall} token dispatch, grouped GEMM, MoE permute fusion, sigmoid router scores with expert bias and sequence-level auxiliary load balancing with coefficient 1e-4. In parallel with the routed experts, each MoE layer also includes an ungated shared expert MLP whose output is added to the combined routed-expert output. 

\begin{table}[!t]
  \caption{Model configs. The final column reports the LatentMoE expert input/output dimension; -- indicates no latent projection.}
  \label{tab:experimental-model-family}
  \centering
  \small
  \resizebox{0.85\linewidth}{!}{%
  \begin{tabular}{llllrrrrr}
    \toprule
    \textbf{Scale} & \textbf{Tokens} & \textbf{Model} & \textbf{Layers} & \textbf{$H$} & \textbf{$|\mathcal{E}|/K$} & \textbf{$f_{\mathrm{expert}}$} & \textbf{$f_{\mathrm{dense}}$} & \textbf{$d_{\mathrm{latent}}$} \\
    \midrule
    2B-A0.5B & 140B & Baseline & \texttt{12M+3A+0D+12E} & 1024 & 128/6 & 640 & -- & -- \\
             &  & CE-MoE   & \texttt{16M+3A+4D+4E}  & 1024 & 160/7 & 1520 & 1920 & -- \\
    \addlinespace
    4B-A0.77B & 215B & Baseline & \texttt{15M+3A+0D+15E} & 1280 & 128/6 & 768 & -- & -- \\
              &  & CE-MoE   & \texttt{20M+3A+5D+5E}  & 1280 & 192/7 & 1504 & 3168 & -- \\
    \addlinespace
    7B-A1B & 310B & Baseline & \texttt{16M+4A+0D+16E} & 1536 & 128/6 & 1024 & -- & -- \\
           &  & CE-MoE   & \texttt{20M+4A+6D+6E}  & 1536 & 192/7 & 1792 & 4576 & -- \\
    \addlinespace
    14B-A2B & 560B & Baseline & \texttt{20M+4A+0D+20E} & 2048 & 128/6 & 1280 & -- & -- \\
            &  & CE-MoE   & \texttt{26M+4A+7D+7E}  & 2048 & 192/7 & 2432 & 3840 & -- \\
    \addlinespace
    31.5B-A3.5B & 300B; 1T & Baseline & \texttt{23M+6A+0D+23E} & 2688 & 128/6 & 1856 & -- & -- \\
                &  & CE-MoE   & \texttt{30M+6A+8D+8E}  & 2688 & 192/7 & 3520 & 8096 & -- \\
 & & LatentMoE & \texttt{23M+6A+0D+23E} & 2688 & 512/22 & 1856 & -- & 672 \\
                &  & CE-LatentMoE & \texttt{30M+6A+8D+8E} & 2688 & 640/24 & 3712 & 6336 & 768 \\
    \bottomrule
  \end{tabular}%
  }
\end{table}

\subsection{Scaling ladder experiments}

We first compare the baseline and CE-MoE under the same global batch size and otherwise matched training configuration. This setting isolates the architecture change and spans five scales from 2B-A0.5B to 31.5B-A3.5B with the scaling-ladder token budgets in Table~\ref{tab:experimental-model-family}.

\begin{figure}[!t]
  \centering
  \includegraphics[width=\linewidth]{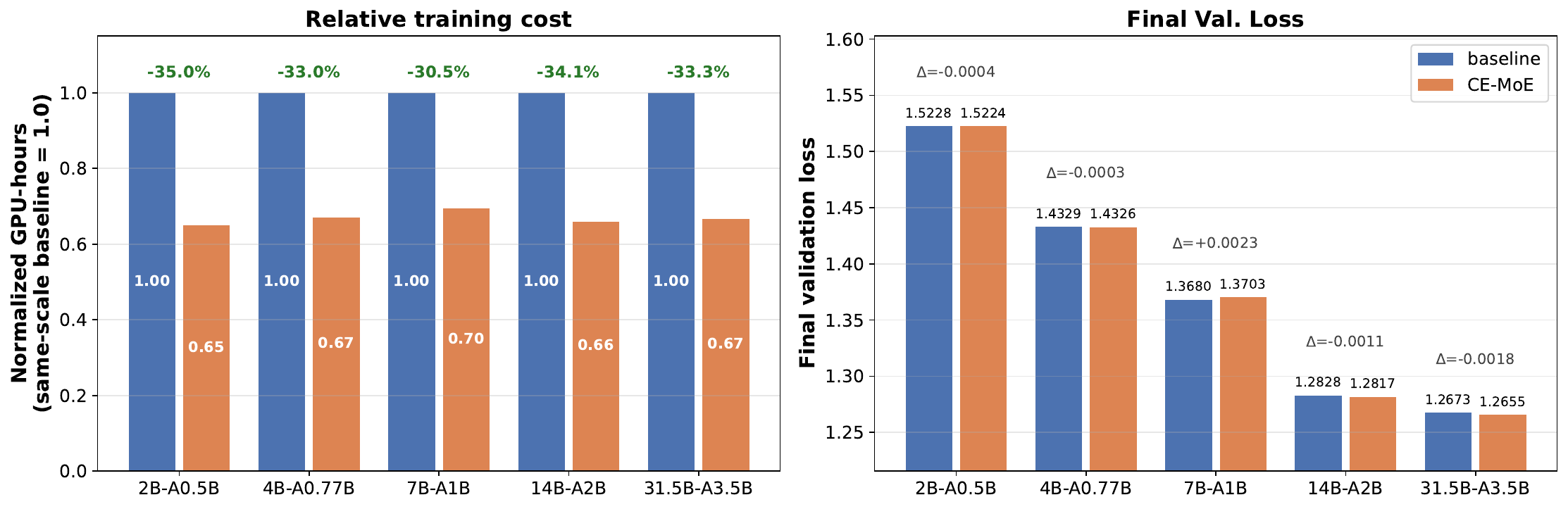}
  \caption{Same-\(\GBS\) scaling ladder up to 31.5B-A3.5B. Training cost is reported as normalized GPU-hours relative to the same-scale baseline. CE-MoE reduces steady-state GPU-hours by 30.5--35.0\% while keeping final validation loss within a few thousandths of the baseline.}
  \label{fig:scaling-ladder-cost-loss}
\end{figure}

Figure~\ref{fig:scaling-ladder-cost-loss} shows that CE-MoE consistently reduces normalized training cost while preserving validation loss. Across the five scales, the GPU-hour reduction ranges from 30.5\% to 35.0\%, while final validation loss stays close to the baseline. This same-\(\GBS\) setting is conservative for CE-MoE: the expert count increases up to 1.5$\times$ the baseline, while the routed top-$K$ increases up to 1.1$\times$ the baseline. Under uniform routing, each CE-MoE expert therefore receives 22\% fewer routed tokens per step despite having more parameters. The fact that CE-MoE remains loss-matched in this regime suggests that the heterogeneous layer pattern contributes useful modeling capacity beyond the parameter-matching procedure itself.

\subsection{Model quality and scaling-up experiments}

Because the same-\(\GBS\) setting gives CE-MoE fewer tokens per expert, we also study whether the saved training cost can be reinvested into a larger global batch. We increase \(\GBS\) to 960 for selected 14B-A2B and 31.5B-A3.5B experiments. This increases the amount of data communicated per iteration, but CE-MoE still remains cheaper than the corresponding baseline while improving downstream quality. With the larger \(\GBS\), we use same parallelism configuration as the baseline while changing the world size. The large-GBS runs are then estimated by averaging runs that use node counts closest to the baseline node count.

Table~\ref{tab:scaling-up-cost-quality} shows that at 14B-A2B, the same-token CE-MoE run uses 34.1\% fewer GPU-hours than the baseline and improves average downstream score from 56.93 to 58.41. The larger-\(\GBS\) CE-MoE run consumes 1.25$\times$ as many tokens, further improves the average to 59.36, and still uses 26.1\% fewer GPU-hours than the baseline. At 31.5B-A3.5B, same-token CE-MoE reduces GPU-hours by 33.3\% while improving the average from 58.37 to 58.65. In the LatentMoE comparison, CE-LatentMoE consumes 1.25$\times$ as many tokens, improves the average from 60.77 to 62.36, and reduces cost by 20.9\% relative to LatentMoE. In the 1T-token long-horizon setting, CE-MoE trains on 1.25$\times$ tokens with 21.0\% fewer GPU-hours and improves the average from 66.26 to 66.67. Overall, these results indicate that communication savings can be traded for either lower training cost at fixed tokens or additional training tokens at still lower cost.

\begin{table}[!t]
  \caption{Training cost and model downstream quality. We report matched-token comparisons for 14B-A2B (560B tokens) and 31.5B-A3.5B (300B tokens), plus selected \(\GBS=960\) CE-MoE/CE-LatentMoE variants trained on 1.25$\times$ tokens. Normalized GPU-hours and GPU-hour reductions are relative to the corresponding baseline in each group. For long-horizon 1T runs, we scale node count to 1.5$\times$ of the 300B token run. \texttt{LatMoE} indicates LatentMoE.}
  \label{tab:scaling-up-cost-quality}
  \centering
  \scriptsize
  \setlength{\tabcolsep}{2.5pt}
  \resizebox{\linewidth}{!}{%
  \begin{tabular}{lrrrrrrrrr}
    \toprule
    & \multicolumn{3}{c}{\textbf{14B-A2B}} & \multicolumn{2}{c}{\textbf{31.5B-A3.5B }} & \multicolumn{2}{c}{\textbf{31.5B-A3.5B }} & \multicolumn{2}{c}{\textbf{31.5B-A3.5B}} \\
    \cmidrule(lr){2-4}\cmidrule(lr){5-6}\cmidrule(lr){7-8}\cmidrule(lr){9-10}
    Model & \texttt{Baseline} & \texttt{CE-MoE} & \texttt{CE-MoE} & \texttt{Baseline} & \texttt{CE-MoE} & \texttt{LatMoE} & \texttt{CE-LatMoE} & \texttt{Baseline} & \texttt{CE-MoE} \\
    Consumed tokens & \texttt{560B} & \texttt{560B} & \texttt{700B} & \texttt{300B} & \texttt{300B} & \texttt{300B} & \texttt{375B} & \texttt{1T} & \texttt{1.25T} \\
    \midrule
    \multicolumn{10}{l}{\textit{Training cost}} \\
    \hspace{0.5em}Norm. GPU-h & 1.00 & 0.66 & 0.74 & 1.00 & 0.67 & 1.00 & 0.79 & 1.00 & 0.79 \\
    \hspace{0.5em}GPU-h reduction & -- & 34.1\% & 26.1\% & -- & 33.3\% & -- & 20.9\% & -- & 21.0\% \\
    \midrule
    \multicolumn{10}{l}{\textit{Downstream quality}} \\
    \hspace{0.5em}MMLU-Pro CoT 5-shot & 29.26 & 30.76 & 33.08 & 30.17 & 31.92 & 35.25 & 37.41 & 42.06 & 44.56 \\
    \hspace{0.5em}MMLU 5-shot & 59.53 & 60.38 & 60.32 & 58.83 & 58.05 & 61.27 & 62.90 & 65.97 & 67.19 \\
    \hspace{0.5em}HumanEval & 41.46 & 39.21 & 42.16 & 42.44 & 42.29 & 43.54 & 45.85 & 47.59 & 49.15 \\
    \hspace{0.5em}MBPP & 53.33 & 53.11 & 55.39 & 55.41 & 54.14 & 57.39 & 58.23 & 61.03 & 62.04 \\
    \hspace{0.5em}GSM8K CoT 8-shot & 64.44 & 69.75 & 73.24 & 67.63 & 69.83 & 72.71 & 71.87 & 81.20 & 78.47 \\
    \hspace{0.5em}MATH-500 & 55.00 & 60.80 & 60.00 & 60.40 & 59.40 & 59.2 & 66.00 & 66.60 & 70.80 \\
    \hspace{0.5em}MATH-Hard & 31.32 & 35.64 & 34.76 & 35.22 & 36.10 & 36.96 & 42.50 & 47.25 & 48.24 \\
    \hspace{0.5em}RACE & 74.16 & 73.88 & 73.88 & 73.78 & 74.74 & 76.08 & 76.08 & 80.96 & 78.76 \\
    \hspace{0.5em}ARC-Challenge & 72.70 & 74.57 & 74.49 & 72.53 & 73.04 & 77.99 & 77.39 & 83.11 & 84.30 \\
    \hspace{0.5em}HellaSwag & 76.11 & 75.90 & 76.52 & 76.94 & 77.01 & 77.73 & 78.09 & 79.46 & 79.56 \\
    \hspace{0.5em}WinoGrande & 68.90 & 68.51 & 69.14 & 68.67 & 68.67 & 70.40 & 69.69 & 73.64 & 70.32 \\
    \midrule
    \textbf{Average} & 56.93 & 58.41 & \textbf{59.36} & 58.37 & \textbf{58.65} & 60.77 & \textbf{62.36} & 66.26 & \textbf{66.67} \\
    \bottomrule
  \end{tabular}%
  }
\end{table}

\subsection{Inference speed}

\begin{figure}[!t]
  \centering
  \includegraphics[width=0.9\linewidth]{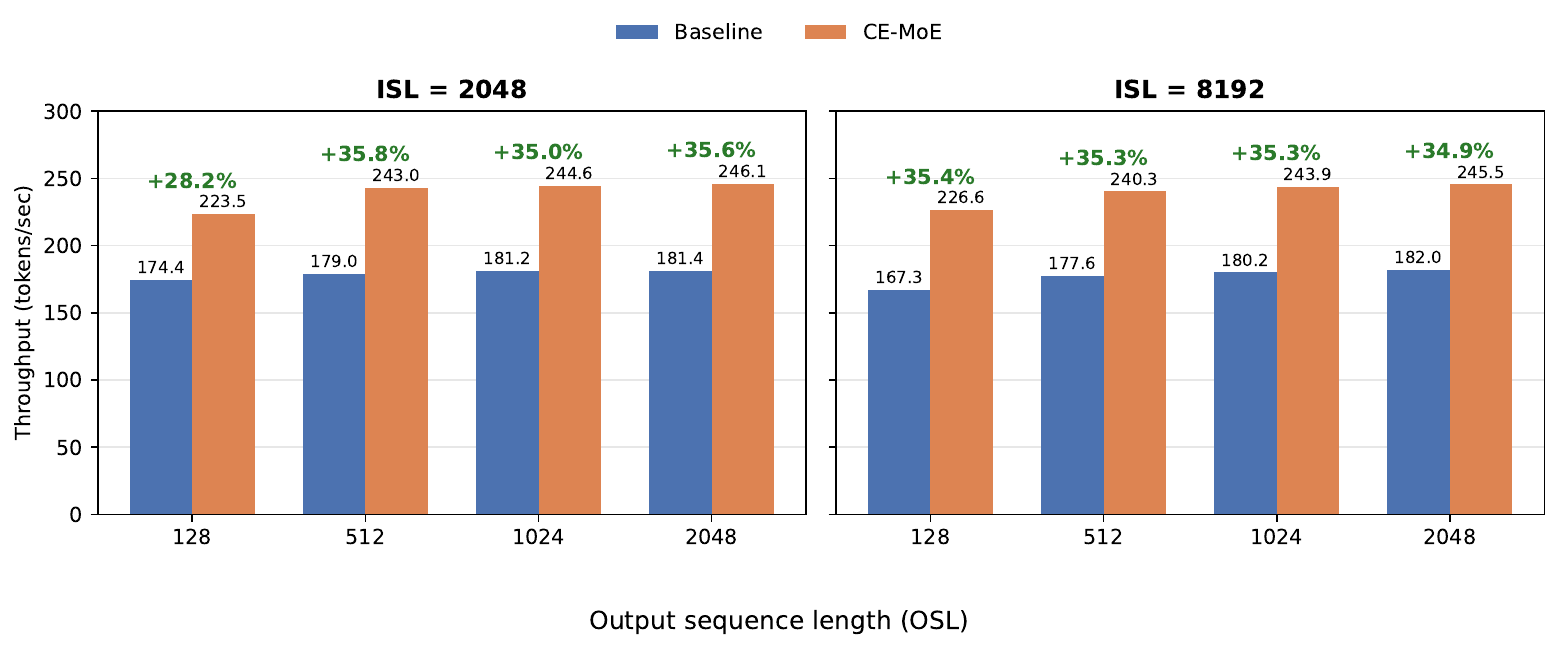}
  \caption{Inference throughput for the 31.5B-A3.5B baseline and CE-MoE at input sequence lengths 2048 and 8192 for various output sequence lengths.}
  \label{fig:inference-speed}
\end{figure}

We run preliminary static, synthetic inference benchmarks using Megatron Server on H100 nodes. The runs use batch size 4, discard two warmup iterations, and report the median over five measured runs. Inference time optimizations are not fully applied here; these results should be viewed as an unoptimized serving proxy rather than a fully tuned deployment. Figure~\ref{fig:inference-speed} reports tokens per second for the 31.5B-A3.5B baseline and the corresponding CE-MoE at input sequence lengths 2048 and 8192 in a single-node setting. Across output sequence lengths from 128 to 2048, CE-MoE improves throughput by 28--36\% over the baseline. Although CE-MoE uses wider expert FFNs, it routes through fewer MoE layers. This reduces router overhead and synchronization points per generated token, which helps explain its higher inference throughput.

\subsection{Residual stream analysis}

\begin{figure}[!h]
  \centering
  \begin{minipage}{0.45\linewidth}
    \centering
    \includegraphics[width=\linewidth]{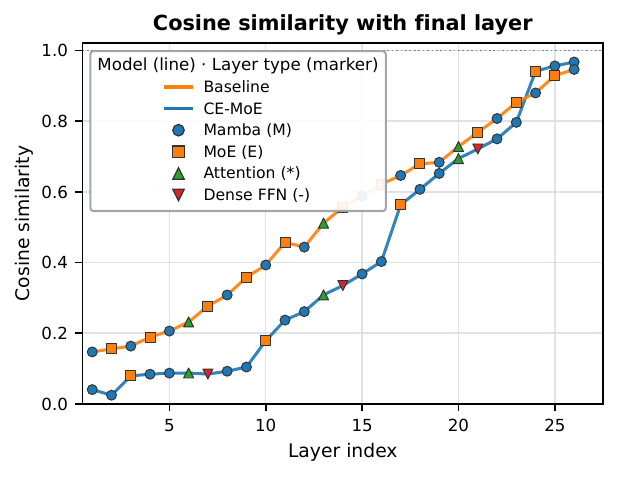}
    \vspace{-0.5em}
    {\small (a) Cosine similarity with final layer}
  \end{minipage}
  \hfill
  \begin{minipage}{0.45\linewidth}
    \centering
    \includegraphics[width=\linewidth]{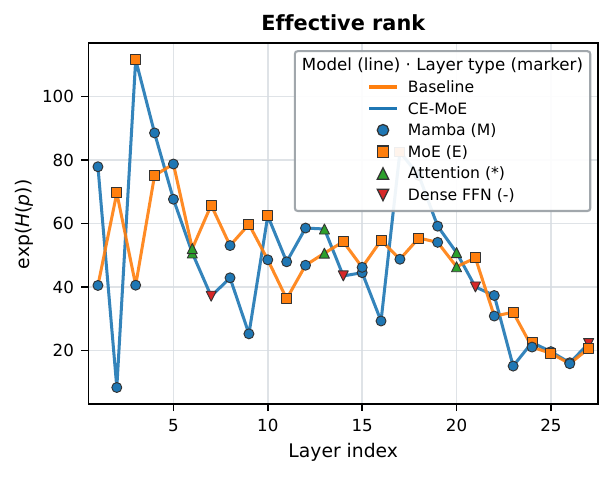}
    \vspace{-0.5em}
    {\small (b) Effective rank}
  \end{minipage}
  \caption{Residual-stream analysis for the 2B-A0.5B baseline and CE-MoE models. Lines identify the model, while marker shapes identify the layer type in each architecture.}
  \label{fig:residual-stream-analysis}
\end{figure}

We further compare the layerwise post-residual representations of the 2B-A0.5B baseline and CE-MoE models at the same training step. For each layer output \(X_\ell\), we measure the per-token cosine similarity to the final-layer residual stream \(X_L\), averaged over tokens, and the effective rank of \(X_\ell\). Figure~\ref{fig:residual-stream-analysis} shows that the baseline's cosine-to-final curve increases relatively smoothly, while CE-MoE follows a more staircase-like trajectory: fewer expert layers make larger representational updates, and the token-mixing layers between expert layers provide smaller refinements.

The effective-rank curves provide a complementary check on this trajectory. Although CE-MoE has larger layer-to-layer rank variation, the two models end with similar effective rank. Rank drops in spans dominated by token mixing but recovers near expert layers, suggesting that these spans do not cause irreversible rank collapse. Together with the cosine pattern, this indicates that CE-MoE reaches comparable validation loss through a different residual-stream trajectory: sparse channel-mixing blocks periodically reshape the representation, while token-mixing layers between them refine it without permanently reducing token diversity.

\section{Conclusion}
\label{sec:limitations-summary}

\paragraph{Summary.}
In this report, we presented communication-efficient hybrid-MoE, which lowers expert-parallel communication by reducing the number of MoE layers \(L_E\) and reallocating capacity into additional token-mixing layers and fewer, wider expert and dense-FFN blocks under comparable total and activated parameter budgets. Across our scaling-ladder experiments, CE-MoE consistently reduces training cost in terms of GPU-hours while keeping validation loss close to matched full-MoE baselines; at larger scale it can train for more tokens at lower cost, remain competitive on downstream benchmarks, and show preliminary inference throughput gains over the full-MoE baseline.

\paragraph{Limitations.}
The measured GPU-hour savings from lowering \(L_E\) may change under other compute or communication precisions (e.g., FP8 or NVFP4), using larger world sizes, or overlapping communication strategies. We also observed that CE-MoE exhibits higher sequence-level load-imbalance signals than full-MoE baselines and suffers from early-training loss spikes more if global batch size does not scale accordingly with expert counts; better routing regularization or load-balancing may improve the efficiency-quality tradeoff. Finally, the inference measurements are preliminary and based on a synthetic static benchmark rather than a full production serving study.

\bibliographystyle{plainnat}
\bibliography{references}

@article{cai2024shortcut,
  author = {Cai, Weilin and Jiang, Juyong and Qin, Le and Cui, Junwei and Kim, Sunghun and Huang, Jiayi},
  title = {{Shortcut-connected expert parallelism for accelerating mixture-of-experts}},
  journal = {arXiv preprint arXiv:2404.05019},
  year = {2024},
  url = {https://arxiv.org/abs/2404.05019}
}

@article{fedus2022switch,
  author = {Fedus, William and Zoph, Barret and Shazeer, Noam},
  title = {{Switch Transformers: Scaling to trillion parameter models with simple and efficient sparsity}},
  journal = {Journal of Machine Learning Research},
  volume = {23},
  number = {120},
  pages = {1--39},
  year = {2022},
  url = {https://arxiv.org/abs/2101.03961}
}

@article{cui2026multiheadlatentmoe,
  author = {Cui, Chenwei and Jackson, Rockwell and Herrera, Benjamin Joseph and T{\'a}rano, Ana Mar{\'i}a and Kerner, Hannah},
  title = {{Multi-head LatentMoE and head parallel: Communication-efficient and deterministic MoE parallelism}},
  journal = {arXiv preprint arXiv:2602.04870},
  year = {2026},
  url = {https://arxiv.org/abs/2602.04870}
}

@misc{deepep,
  author = {{DeepSeek-AI}},
  title = {{DeepEP: An efficient expert-parallel communication library}},
  howpublished = {GitHub repository},
  year = {2025},
  url = {https://github.com/deepseek-ai/DeepEP}
}

@article{dao2024transformers,
  author = {Dao, Tri and Gu, Albert},
  title = {{Transformers are SSMs: Generalized Models and Efficient Algorithms Through Structured State Space Duality}},
  journal = {arXiv preprint arXiv:2405.21060},
  year = {2024},
  url = {https://arxiv.org/abs/2405.21060}
}

@inproceedings{dong2021attention,
  author = {Dong, Yihe and Cordonnier, Jean-Baptiste and Loukas, Andreas},
  title = {{Attention is not all you need: Pure attention loses rank doubly exponentially with depth}},
  booktitle = {Proceedings of the 38th International Conference on Machine Learning},
  series = {Proceedings of Machine Learning Research},
  volume = {139},
  pages = {2793--2803},
  year = {2021},
  url = {https://proceedings.mlr.press/v139/dong21a.html}
}

@article{elango2026latentmoe,
  author = {Elango, Venmugil and Bhatia, Nidhi and Waleffe, Roger and Shafipour, Rasoul and Asida, Tomer and Khattar, Abhinav and Assaf, Nave and Golub, Maximilian and Guman, Joey and Mitra, Tiyasa and Zhao, Ritchie and Borkar, Ritika and Zilberstein, Ran and Patwary, Mostofa and Shoeybi, Mohammad and Rouhani, Bita},
  title = {{LatentMoE: Toward optimal accuracy per FLOP and parameter in mixture of experts}},
  journal = {arXiv preprint arXiv:2601.18089},
  year = {2026},
  url = {https://arxiv.org/abs/2601.18089}
}

@article{guo2025sonicmoe,
  author = {Guo, Wentao and Mishra, Mayank and Cheng, Xinle and Stoica, Ion and Dao, Tri},
  title = {{SonicMoE: Accelerating MoE with IO and tile-aware optimizations}},
  journal = {arXiv preprint arXiv:2512.14080},
  year = {2025},
  url = {https://arxiv.org/abs/2512.14080}
}

@inproceedings{jin2026megascale,
  author = {Jin, Chao and Jiang, Ziheng and Bai, Zhihao and Zhong, Zheng and Liu, Juncai and Li, Xiang and Zheng, Ningxin and Wang, Xi and Xie, Cong and Huang, Qi and Heng, Wen and Ma, Yiyuan and Bao, Wenlei and Zheng, Size and Zheng, Xuegui and Peng, Yanghua and Lin, Haibin and Liu, Xuanzhe and Jin, Xin and Liu, Xin},
  title = {{MegaScale-MoE: Large-Scale Communication-Efficient Training of Mixture-of-Experts Models in Production}},
  booktitle = {Proceedings of the 21st European Conference on Computer Systems},
  pages = {366--382},
  publisher = {ACM},
  year = {2026},
  doi = {10.1145/3767295.3769325},
  url = {https://dl.acm.org/doi/10.1145/3767295.3769325}
}

@article{lieber2024jamba,
  author = {Lieber, Opher and Lenz, Barak and Bata, Hofit and Cohen, Gal and Osin, Jhonathan and Dalmedigos, Itay and Safahi, Erez and Meirom, Shaked and Belinkov, Yonatan and Shalev-Shwartz, Shai and Abend, Omri and Alon, Raz and Asida, Tomer and Bergman, Amir and Glozman, Roman and Gokhman, Michael and Manevich, Avashalom and Ratner, Nir and Rozen, Noam and Shwartz, Erez and Zusman, Mor and Shoham, Yoav},
  title = {{Jamba: A Hybrid Transformer-Mamba Language Model}},
  journal = {arXiv preprint arXiv:2403.19887},
  year = {2024},
  url = {https://arxiv.org/abs/2403.19887}
}

@article{lepikhin2020gshard,
  author = {Lepikhin, Dmitry and Lee, HyoukJoong and Xu, Yuanzhong and Chen, Dehao and Firat, Orhan and Huang, Yanping and Krikun, Maxim and Shazeer, Noam and Chen, Zhifeng},
  title = {{GShard: Scaling giant models with conditional computation and automatic sharding}},
  journal = {arXiv preprint arXiv:2006.16668},
  year = {2020},
  url = {https://arxiv.org/abs/2006.16668}
}

@article{loshchilov2024ngpt,
  author = {Loshchilov, Ilya and Hsieh, Cheng-Ping and Sun, Simeng and Ginsburg, Boris},
  title = {{nGPT: Normalized Transformer with representation learning on the hypersphere}},
  journal = {arXiv preprint arXiv:2410.01131},
  year = {2024},
  url = {https://arxiv.org/abs/2410.01131}
}

@techreport{mai2026thinking1,
  author = {{The Microsoft AI Team}},
  title = {{MAI-Thinking-1: Building a Hill-Climbing Machine}},
  institution = {Microsoft AI},
  year = {2026},
  url = {https://microsoft.ai/pdf/mai-thinking-1.pdf}
}

@article{nvidia2025nvfp4,
  author = {{NVIDIA} and others},
  title = {{Pretraining Large Language Models with NVFP4}},
  journal = {arXiv preprint arXiv:2509.25149},
  year = {2025},
  url = {https://arxiv.org/abs/2509.25149}
}

@article{nvidia2025nemotron3,
  author = {{NVIDIA} and others},
  title = {{NVIDIA Nemotron 3: Efficient and Open Intelligence}},
  journal = {arXiv preprint arXiv:2512.20856},
  year = {2025},
  url = {https://arxiv.org/abs/2512.20856}
}

@article{nvidia2026nemotron3super,
  author = {{NVIDIA} and others},
  title = {{Nemotron 3 Super: Open, Efficient Mixture-of-Experts Hybrid Mamba-Transformer Model for Agentic Reasoning}},
  journal = {arXiv preprint arXiv:2604.12374},
  year = {2026},
  url = {https://arxiv.org/abs/2604.12374}
}

@article{qiu2025gatedattention,
  author = {Qiu, Zihan and Wang, Zekun and Zheng, Bo and Huang, Zeyu and Wen, Kaiyue and Yang, Songlin and Men, Rui and Yu, Le and Huang, Fei and Huang, Suozhi and Liu, Dayiheng and Zhou, Jingren and Lin, Junyang},
  title = {{Gated attention for large language models: Non-linearity, sparsity, and attention-sink-free}},
  journal = {arXiv preprint arXiv:2505.06708},
  year = {2025},
  url = {https://arxiv.org/abs/2505.06708}
}

@article{yan2026megatroncore,
  author = {Yan, Zijie and others},
  title = {{Scalable training of mixture-of-experts models with Megatron Core}},
  journal = {arXiv preprint arXiv:2603.07685},
  year = {2026},
  url = {https://arxiv.org/abs/2603.07685}
}

@article{zoph2022stmoe,
  author = {Zoph, Barret and Bello, Irwan and Kumar, Sameer and Du, Nan and Huang, Yanping and Dean, Jeff and Shazeer, Noam and Fedus, William},
  title = {{ST-MoE: Designing stable and transferable sparse expert models}},
  journal = {arXiv preprint arXiv:2202.08906},
  year = {2022},
  url = {https://arxiv.org/abs/2202.08906}
}

@inproceedings{hendrycks2021mmlu,
  author = {Hendrycks, Dan and Burns, Collin and Basart, Steven and Zou, Andy and Mazeika, Mantas and Song, Dawn and Steinhardt, Jacob},
  title = {{Measuring Massive Multitask Language Understanding}},
  booktitle = {International Conference on Learning Representations},
  year = {2021},
  url = {https://arxiv.org/abs/2009.03300}
}

@article{wang2024mmlupro,
  author = {Wang, Yubo and Ma, Xueguang and Zhang, Ge and Ni, Yuansheng and Chandra, Abhranil and Guo, Shiguang and Ren, Weiming and Arulraj, Aaran and He, Xuan and Jiang, Ziyan and Li, Tianle and Ku, Max and Wang, Kai and Zhuang, Alex and Fan, Rongqi and Yue, Xiang and Chen, Wenhu},
  title = {{MMLU-Pro: A More Robust and Challenging Multi-Task Language Understanding Benchmark}},
  journal = {arXiv preprint arXiv:2406.01574},
  year = {2024},
  url = {https://arxiv.org/abs/2406.01574}
}

@article{chen2021evaluating,
  author = {Chen, Mark and others},
  title = {{Evaluating Large Language Models Trained on Code}},
  journal = {arXiv preprint arXiv:2107.03374},
  year = {2021},
  url = {https://arxiv.org/abs/2107.03374}
}

@article{austin2021program,
  author = {Austin, Jacob and Odena, Augustus and Nye, Maxwell and Bosma, Maarten and Michalewski, Henryk and Dohan, David and Jiang, Ellen and Cai, Carrie and Terry, Michael and Le, Quoc and Sutton, Charles},
  title = {{Program Synthesis with Large Language Models}},
  journal = {arXiv preprint arXiv:2108.07732},
  year = {2021},
  url = {https://arxiv.org/abs/2108.07732}
}

@article{cobbe2021training,
  author = {Cobbe, Karl and Kosaraju, Vineet and Bavarian, Mohammad and Chen, Mark and Jun, Heewoo and Kaiser, Lukasz and Plappert, Matthias and Tworek, Jerry and Hilton, Jacob and Nakano, Reiichiro and Hesse, Christopher and Schulman, John},
  title = {{Training Verifiers to Solve Math Word Problems}},
  journal = {arXiv preprint arXiv:2110.14168},
  year = {2021},
  url = {https://arxiv.org/abs/2110.14168}
}

@article{hendrycks2021math,
  author = {Hendrycks, Dan and Burns, Collin and Kadavath, Saurav and Arora, Akul and Basart, Steven and Tang, Eric and Song, Dawn and Steinhardt, Jacob},
  title = {{Measuring Mathematical Problem Solving With the MATH Dataset}},
  journal = {arXiv preprint arXiv:2103.03874},
  year = {2021},
  url = {https://arxiv.org/abs/2103.03874}
}

@article{lightman2023lets,
  author = {Lightman, Hunter and Kosaraju, Vineet and Burda, Yura and Edwards, Harri and Baker, Bowen and Lee, Teddy and Leike, Jan and Schulman, John and Sutskever, Ilya and Cobbe, Karl},
  title = {{Let's Verify Step by Step}},
  journal = {arXiv preprint arXiv:2305.20050},
  year = {2023},
  url = {https://arxiv.org/abs/2305.20050}
}

@article{lewkowycz2022solving,
  author = {Lewkowycz, Aitor and Andreassen, Anders and Dohan, David and Dyer, Ethan and Michalewski, Henryk and Ramasesh, Vinay and Slone, Ambrose and Anil, Cem and Schlag, Imanol and Gutman-Solo, Theo and Wu, Yuhuai and Neyshabur, Behnam and Gur-Ari, Guy and Misra, Vedant},
  title = {{Solving Quantitative Reasoning Problems with Language Models}},
  journal = {arXiv preprint arXiv:2206.14858},
  year = {2022},
  url = {https://arxiv.org/abs/2206.14858}
}

@inproceedings{lai2017race,
  author = {Lai, Guokun and Xie, Qizhe and Liu, Hanxiao and Yang, Yiming and Hovy, Eduard},
  title = {{RACE: Large-scale ReAding Comprehension Dataset From Examinations}},
  booktitle = {Proceedings of the 2017 Conference on Empirical Methods in Natural Language Processing},
  pages = {785--794},
  year = {2017},
  url = {https://arxiv.org/abs/1704.04683}
}

@inproceedings{zellers2019hellaswag,
  author = {Zellers, Rowan and Holtzman, Ari and Bisk, Yonatan and Farhadi, Ali and Choi, Yejin},
  title = {{HellaSwag: Can a Machine Really Finish Your Sentence?}},
  booktitle = {Proceedings of the 57th Annual Meeting of the Association for Computational Linguistics},
  pages = {4791--4800},
  year = {2019},
  url = {https://arxiv.org/abs/1905.07830}
}

@inproceedings{sakaguchi2020winogrande,
  author = {Sakaguchi, Keisuke and Le Bras, Ronan and Bhagavatula, Chandra and Choi, Yejin},
  title = {{WinoGrande: An Adversarial Winograd Schema Challenge at Scale}},
  booktitle = {Proceedings of the AAAI Conference on Artificial Intelligence},
  year = {2020},
  url = {https://arxiv.org/abs/1907.10641}
}

@article{clark2018think,
  author = {Clark, Peter and Cowhey, Isaac and Etzioni, Oren and Khot, Tushar and Sabharwal, Ashish and Schoenick, Carissa and Tafjord, Oyvind},
  title = {{Think you have Solved Question Answering? Try ARC, the AI2 Reasoning Challenge}},
  journal = {arXiv preprint arXiv:1803.05457},
  year = {2018},
  url = {https://arxiv.org/abs/1803.05457}
}

\end{document}